\pdfoutput=1
\documentclass[conference]{IEEEtran}
\IEEEoverridecommandlockouts
\usepackage{cite}
\usepackage{amsmath,amssymb,amsfonts}
\usepackage{algorithmic}
\usepackage{graphicx}
\usepackage{textcomp}
\usepackage{xcolor}
\usepackage[super]{nth}
\usepackage{setspace}
\usepackage{caption}
\usepackage[T1]{fontenc}
\usepackage{subfig}
\usepackage{multirow}
\usepackage{nth}
\usepackage{diagbox}
\usepackage{bm}

\def\BibTeX{{\rm B\kern-.05em{\sc i\kern-.025em b}\kern-.08em
    T\kern-.1667em\lower.7ex\hbox{E}\kern-.125emX}}

\begin{document}

\title{Amplifying the Anterior-Posterior Difference via Data Enhancement - A More Robust Deep Monocular Orientation Estimation Solution}

\author{Chenchen Zhao and Hao Li$^{*}$
\thanks{This research work is supported by the SJTU (Shanghai Jiao Tong Univ.) Young Talent Funding (WF220426002).}
\thanks{Chenchen Zhao is with Dept. Automation, SJTU, Shanghai, 200240, China. }
\thanks{Hao Li, Assoc. Prof., is with Dept. Automation and SPEIT, SJTU, Shanghai, 200240, China. }
\thanks{* Corresponding author: Hao Li ({Email:\tt\small haoli@sjtu.edu.cn})}
}

\maketitle

\begin{abstract}

Existing deep-learning based monocular orientation estimation algorithms faces the problem of
confusion between the anterior and posterior parts of the objects,
caused by the feature similarity of such parts in typical objects in traffic scenes
such as cars and pedestrians.
While difficult to solve, the problem may lead to serious orientation estimation errors,
and pose threats to the upcoming decision making process of the ego vehicle,
since the predicted tracks of objects may have directions opposite to ground truths.
In this paper, we mitigate this problem by proposing a pretraining method.
The method focuses on predicting the left/right semicircle in which the orientation of the object is located.
The trained semicircle prediction model is then integrated into the orientation angle estimation model
which predicts a value in range $[0, \pi]$.
Experiment results show that the proposed semicircle prediction
enhances the accuracy of orientation estimation, and mitigates the problem stated above.
With the proposed method, a backbone achieves similar state-of-the-art orientation estimation performance
to existing approaches with well-designed network structures.

\end{abstract}


\section{Introduction}\label{intro}

Obstacle avoidance is a significant step in autonomous driving and intelligent vehicle design.
In complex traffic environments,
moving objects such as cars and pedestrians are harder to detect and localize,
and are yet more threatening obstacles than stationary objects.
In such environments, perception modules with orientation estimation
are better at satisfying the needs for safety than those with pure object detection.
Orientation estimation modules play the important role of
predicting the heading directions and short-term tracks \cite{4621257} of the surrounding objects,
which serve as references to a set of safety measures determined and carried out by the ego vehicle,
such as path planning, speed control and risk assessment \cite{survey,benchmark,enzweiler2010integrated}.
Therefore, as one fundamental step of track prediction, orientation estimation is a vital ability
of modern perception modules in autonomous vehicles \cite{gd}.

From the characteristics and applications of orientation estimation,
it is more suitable and more important in monocular occasions,
since the information the monocular cameras can obtain is much more scarce than other types of sensors.
With the depth information unknown, orientation becomes the only information for tracking.
From another perspective, monocular solutions are more resource-saving with less complexity,
and are vital in occasions with LiDAR not available or suitable \cite{ffnet}.
Therefore, monocular orientation estimation attracts much research interest in autonomous driving
\cite{ffnet,deep3dbox,4621257,enzweiler2010integrated}.

Since orientation estimation is based on object localization,
solutions to the problem are mostly attached to classic object detection methods.
However, apart from difficulties in object localization, orientation estimation faces several more problems,
with the main ones listed as follows:

First, More and deeper features required: 
Object localization models have the ability of extracting object-level features.
On the contrary, orientation estimation requires more component-level features,
since orientation is estimated with advantage of
relative positions of specific components to the whole object.

Second, Lack of prior information:
There are much less prior knowledge available for orientation estimation than object localization
(e.g. exclusion of obviously irrational locations of cars and pedestrians).
All directions are reasonable for any objects.

Third, Confusion between the anterior and posterior parts of objects:
The anterior and posterior parts have similar component-level features
(e.g. similar light, license plate and window positions, etc),
which may confuse the orientation estimation methods to output results just opposite to the truth.
The problem is more severe in pedestrian orientation estimation,
since the most notable face characteristics cannot be used for orientation estimation
(face orientation is not necessarily equal to body orientation).

Such problems are especially tough for monocular solutions:
First, depth information of objects is not available for them;
second, the anterior and posterior parts of objects cannot both be visible without occlusion
at the same time in monocular images, making the third problem even harder to solve.
In most existing orientation estimation solutions,
object localization and orientation estimation share one feature extractor,
indicating that the features of objects are still scarce in the orientation estimation process.
In the meantime, little existing research focus on the second and the third problem.
Therefore, the above three problems still exist in orientation estimation.

We focus on the third problem stated above in monocular orientation estimation,
and propose a monocular pretraining method to enhance performance of the orientation estimation model,
while in the meantime, partially solving the first and the second problem.
The pretraining method predicts the left/right semicircle in which the orientation of the object is located.
With the pretraining method, the original orientation estimation problem is converted to
estimation of an absolute orientation value in range $[0, \pi]$.
We implement the method in both supervised and semisupervised manners,
and test their performance in typical traffic scenarios.
Experiment results with the KITTI \cite{kitti} dataset show that the proposed pretraining method 
effectively improves the accuracy of orientation estimation.
A backbone model achieves state-of-the-art orientation estimation performance with the proposed method,
similar to existing approaches with well-designed network structures.
Our contributions are summarized as follows:
first, we propose a novel pretraining method in orientation estimation
and achieve state-of-the-art performance;
second, we offer a new perspective of features available for orientation estimation,
with the proposed method being a way of feature mining and augmentation.

The paper is organized as follows:
we introduce some related work in orientation estimation in Section \ref{relatedwork};
introduction and details of the semicircle pretraining method are stated in Section \ref{method};
experiments are conducted in Section \ref{experiment}
to validate the effectiveness and performance of the proposed method;
we conclude the paper in Section \ref{conclusion}.

\addtolength{\topmargin}{-0.25in}
\addtolength{\textheight}{0.25in}

\section{Related Work}\label{relatedwork}

Although with equally much importance,
orientation estimation attracts much less attention than object detection,
according to recent studies and researches in the related field.

The orientation estimation module in most existing monocular methods
\cite{mono3d,subcnn,monopsr,shiftrcnn,ipm,mvra}
is basically a prediction branch connected to the feature extractor of a well-designed object detection model.
Some work pays special attention to orientation estimation and makes related improvements:
In \cite{deep3dbox}, the authors proposed the Multibin mechanism,
to divide the orientation domain into several sectors, converting the problem into
a high-level classification and low-level residual regression;
another discretized method similar to Multibin is proposed in \cite{dcnn};
3D modelling is adopted in \cite{symgan}, and the problem is solved in an unsupervised manner
using GAN and the 3D models;
the authors in \cite{ffnet} introduce the relationship between
2D/3D dimensions and orientation angles of objects to orientation estimation as supplementary prior knowledge.

While having accuracy improvements on object orientation estimation to some extent,
the methods above may not be qualified for complex orientation estimation problems.
As stated above, orientation estimation and object detection require different levels of features,
indicating that one feature extractor may not be qualified for solving both problems.
The same problem happens on the Multibin classifier.
Besides, since object features with different orientations may appear to be rather similar
(e.g. features of exactly opposite orientations),
the sector-classification solution may be confused of feature-similar orientations.
The anterior-posterior-similarity problem still exists in \cite{ffnet}
in which 2D and 3D dimensions are independent from anterior and posterior features.
In the meantime, the method proposed in \cite{symgan} relies heavily on the predetermined 3D model,
thus restricted to simple scenarios with available 3D models of all objects.

In this paper, we propose a method specially intended for orientation estimation.
We focus on the neglected anterior-posterior-similarity problem
and propose a novel pretraining method to solve it.

\section{Prediction of Orientation Semicircle as Pretraining}\label{method}

\subsection{Definition and Choice of Orientation Semicircles}\label{choice}

To strengthen the ability of anterior-posterior classification of the orientation estimation model,
we connect a classification branch to the feature extractor of the original model,
and train the feature extractor and the branch firstly.
The anterior-posterior classification is a coarse pretraining step
yet helpful to better orientation estimation.

\begin{figure}[tb]
\centering
\includegraphics[width=0.6\linewidth]{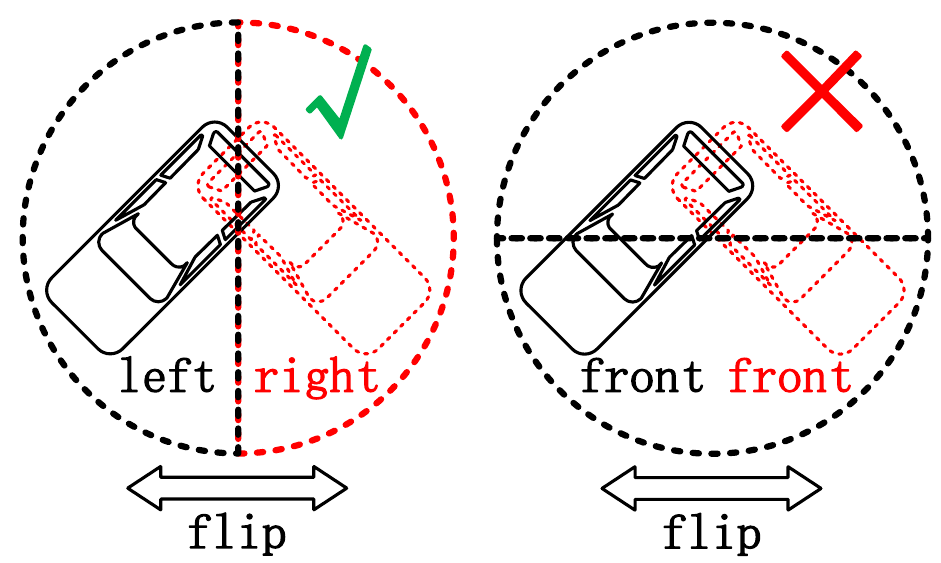}
\caption{Choice of the semicircles and related dataset enhancement.}
\label{semicircle}
\end{figure}

Inspired by the sector classfication spirit of Multibin,
the proposed method first divides the orientation domain into several sectors.
With the aim of amplifying the differences between the anterior and posterior parts,
the number of sectors is set 2 to make the target of the pretraining process
a pure anterior-posterior classification.
In consideration of dataset enhancement,
locations of the two sectors are set left and right instead of front and back.
With the left/right location settings, dataset is enhanced by simple flip of images,
and each object forms a pair of data with different semicircle classification labels.
Choice of the semicircles and related dataset enhancement are shown in Fig. \ref{semicircle}.

\subsection{Supervised Orientation Estimation with Semicircle Prediction as Pretraining}

The semicircle prediction serves as a pretraining step,
and is combined with the main orientation estimation process.

\begin{figure}[tb]
\centering
\includegraphics[width=\linewidth]{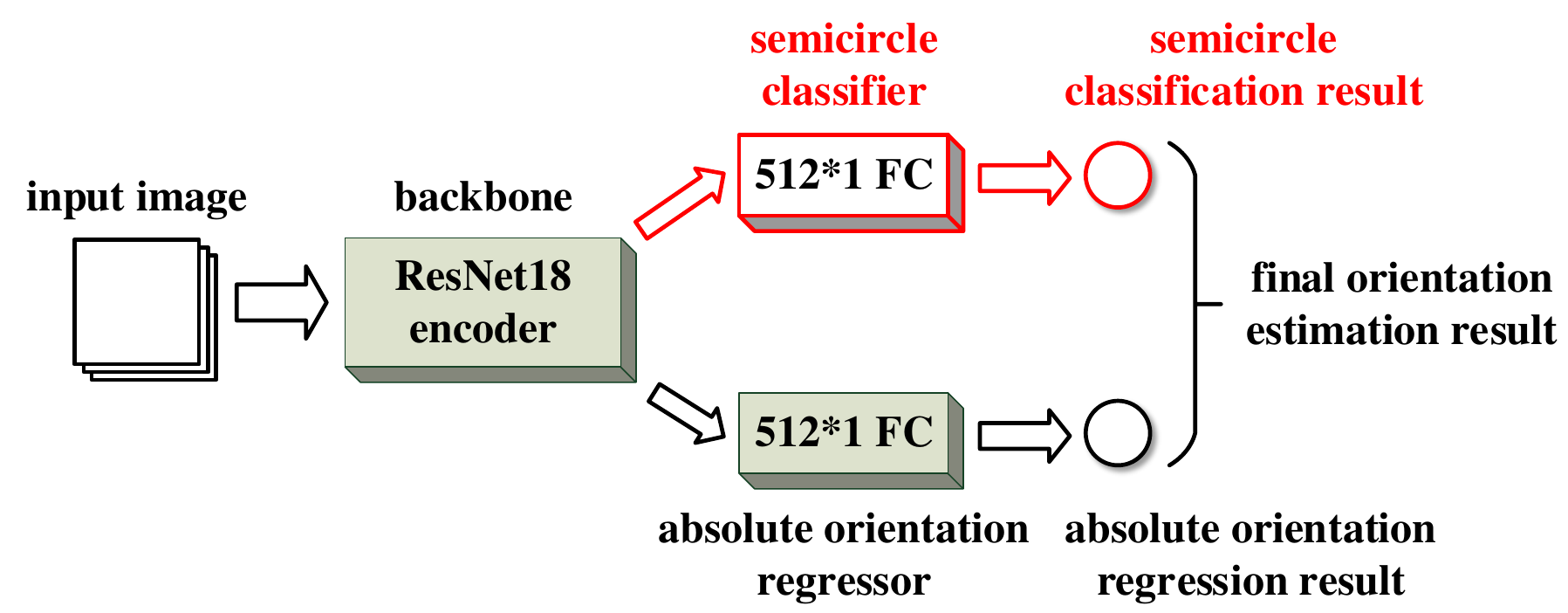}
\caption{Network structure of the proposed orientation estimation model.}
\label{structure}
\end{figure}

The network structure of the whole orientation estimation model is shown in Fig. \ref{structure}.
The network adopts the ResNet18 \cite{resnet} backbone, and has one classification and one regression branch.
The classification branch is stated in Section \ref{choice},
while the regression branch outputs the absolute orientation value in range $[0, \pi]$.
The two results jointly determine the estimated orientation, which is shown in \eqref{joint}.
\begin{equation}
\begin{aligned}
\alpha = (i_{C(F(x))} - 1)\pi + R(F(x))
\end{aligned}
\label{joint}
\end{equation}
in which $F$, $C$ and $R$ are respectively the feature extractor, the classifier and the regressor;
$x$ is the image input corresponding to only one object;
$i_{C(F(x))}$ is the index of the classification result; $R(F(x))$ is the regression result.
Note that the regressor has no additional designs for orientation estimation.

We decompose the whole process into three steps for better and more robust training.
In the first step, the classifier and the feature extractor are jointly trained
while parameters of the regressor are frozen.
In this step, loss is defined in \eqref{loss1}:
\begin{equation}
\begin{aligned}
L_1 = L_{CE}(C(F(x)), \epsilon(y))
\end{aligned}
\label{loss1}
\end{equation}
in which $y$ is the orientation label of $x$;
$L_{CE}$ is the classic cross entropy criterion in classification; $\epsilon$ is the step function.
In this step, a qualified semicircle classifier is established, while in the meantime,
a feature extractor capable of extracting basic orientation features is also trained.
The feature extractor is trained to focus more on component-level features.

In the second step, the regressor and the feature extractor are jointly trained
while parameters of the classifier are frozen.
In this step, loss is defined in \eqref{loss2}:
\begin{equation}
\begin{aligned}
L_2 = L_1 + L_{MSE}(R(F(x)), \cos{(y + (\epsilon(y) - 1)\pi)})
\end{aligned}
\label{loss2}
\end{equation}
in which $L_{MSE}$ is the mean squared error criterion in regression.
In this step, a qualified orientation estimator is established, while in the meantime,
the feature extractor is enhanced, and can extract more precise orientation features.

In the final step, all modules of the model are tuned simultaneously. Loss $L_2$ is adopted.

Details of the three steps are shown in Table \ref{detail}.

\begin{table}[tb]
\centering
\caption{Details of the three steps in the proposed method.}
\begin{tabular}{|c|c|c|c|c|c|}
\hline
step & F & C & R & $loss_{CE}$ & $loss_{MSE}$ \\
\hline
1 & \checkmark & \checkmark & & \checkmark & \\
\hline
2 & \checkmark & & \checkmark & \checkmark & \checkmark \\
\hline
3 & \checkmark & \checkmark & \checkmark & \checkmark & \checkmark \\
\hline
\end{tabular}
\label{detail}
\end{table}

The proposed method solves the third problem and partially solves the first and second problems
stated in Section \ref{intro} with the following reasons:

\begin{itemize}
\item (to problem 1)
Features of orientation semicircles are closely related to the final orientation results,
thus serving as supplementary information in orientation estimation.
\item (to problem 2)
The following prior information is introduced to orientation estimation:
the semicircle label of an image is opposite to that of the corresponding horizontal-flipped image.
\item (to problem 3)
The semicircle classification has no essential difference from the anterior-posterior classification.
An effective solution to the former problem will be equally effective to the latter one.
\end{itemize}

\subsection{Semisupervised Orientation Estimation with Unsupervised Semicircle Prediction}

The method can also be implemented in a semisupervised way with the first step unsupervised.

As stated in Section \ref{choice},
each object forms a pair of data with different semicircle classification labels
with a simple horizontal flip of the image.
Choices of locations of the semicircles as left and right enable not only dataset enhancement,
but also unsupervised training,
since the unsupervised criterion can be set
the difference between prediction results of the elements in each pair.
The unsupervised criterion is shown in \eqref{unsupervisedloss},
in which $x_1$ and $x_2$ are respectively the original and flipped image:
\begin{equation}
\begin{aligned}
L_u = L_{CE}(C(F(x_1)), \overline{C(F(x_2))})
\end{aligned}
\label{unsupervisedloss}
\end{equation}
$C(F(x_1))$ (result of the original image) and $C(F(x_2))$ (result of the flipped image)
are `considered' the output and the label of the semicircle classification
(although the correctness of them remains undetermined).

With the randomly decided semicircle label in training,
the semicircle classifier may not be absolutely accurate.
As a result, parameters of the classifier will not be frozen,
and the second step is deleted from semisupervised training.
Training details of the third step are similar to the supervised training frame.

\section{Experiments}\label{experiment}

Both supervised and semisupervised training are experimented with results and analyses.
Experimental comparisons with the plain model and with state-of-the-art baselines are conducted.

\subsection{Experiment Setup}

The dataset of KITTI Object Detection and Orientation Estimation Evaluation benchmark \cite{kitti} is used.
We conduct orientation estimation experiments on both cars and pedestrians.
Hyperparameters in training are shown in Table \ref{params}.

\begin{table}[]
\centering
\caption{Hyperparameters in supervised and semisupervised training in the proposed method.}
\begin{tabular}{|c|c|c|c|}
\hline
\multirow{2}*{Hyperparameter} & \multicolumn{3}{c|}{Value}\\
\cline{2-4}
 & step1 & step2 & step3 \\
\hline
image size & \multicolumn{3}{c|}{$224\times 224$} \\
\hline
learning rate & $10^{-5}$ & $10^{-5}$ & $5\times 10^{-6}$ \\
\hline
number of iterations & 250K & 150K & 100K \\
\hline
batch size & \multicolumn{3}{c|}{16} \\
\hline
\end{tabular}
\label{params}
\end{table}

\subsection{Results of Vanilla Training on Backbone}

We first conduct vanilla training on the backbone to prove that the anterior-posterior problem does exist.

\begin{figure}[tb]
\centering
\includegraphics[width=0.6\linewidth]{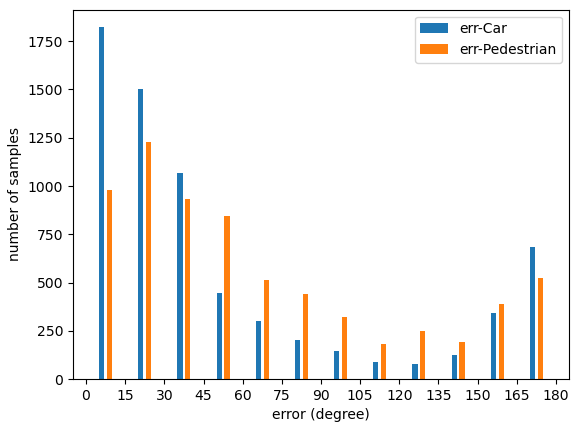}
\caption{The mean squared orientation error of the plain model.}
\label{vanilla}
\end{figure}

After the identical training process without the semicircle classifier,
the backbone has the mean squared orientation error shown in Fig. \ref{vanilla}.
From the experiment results, it is obvious that
the backbone is weak at classifying the anterior and posterior parts of objects,
resulting in the two error peaks (0 and $\pi$) shown in Fig. \ref{vanilla}.
Therefore, the anterior-posterior problem is an existing urgent problem.

\subsection{Results of Supervised Training on Backbone with Semicircle Classifier}

\begin{figure}[tb]
\centering
\includegraphics[width=0.6\linewidth]{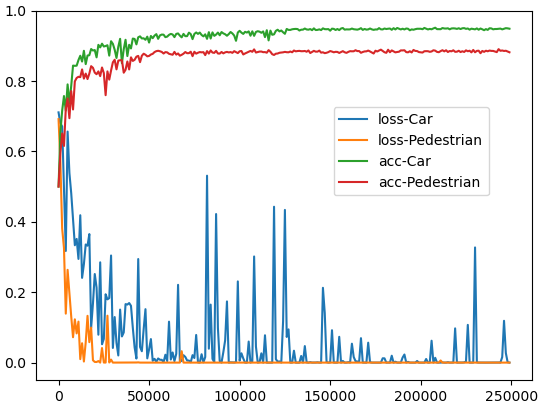}
\caption{The loss convergence and accuracy increase of the semicircle classifier in the first step.}
\label{sup_step1}
\end{figure}

In training of the first step, the loss convergence and accuracy increase of the semicircle classifier
with both car and pedestrian objects are shown in Fig. \ref{sup_step1}.
From the results, it is obvious that the semicircle classification method is effective and accurate.
The classifier reaches about 95\% and 90\% accuracy on cars and pedestrians.

\begin{figure}[tb]
\centering
\subfloat[The loss convergence of the semicircle classifier.]{
\includegraphics[width=0.45\linewidth]{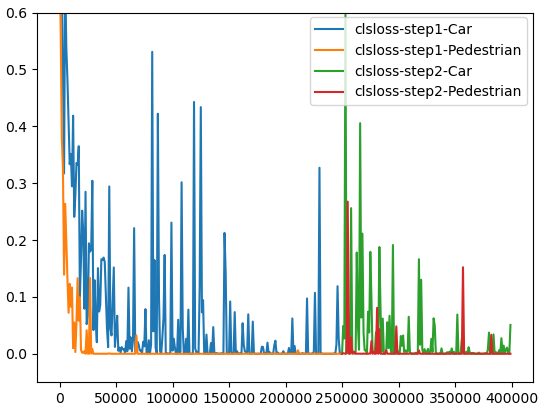}
\label{sup_step2_cls}
}
\subfloat[The loss convergence of the orientation regressor and error decrease of the joint model.]{
\includegraphics[width=0.45\linewidth]{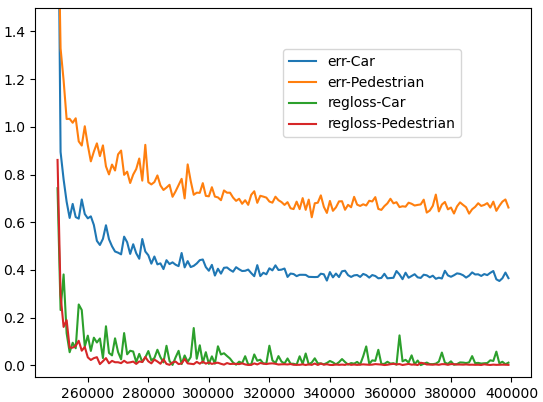}
\label{sup_step2_reg_err}
}
\caption{The loss convergence and error decrease in the second step.}
\end{figure}

In training of the second step, the classification loss of the semicircle classifier
is shown in Fig. \ref{sup_step2_cls}.
Although facing slight oscillations,
loss of the classifier converges to a satisfactory level with a fast speed.
At the beginning of the second step, the loss value is much smaller than
that at the beginning of the first step, indicating the effectiveness of the pretraining method.
The loss convergence and error decrease of the orientation regressor
are shown in Fig. \ref{sup_step2_reg_err}.
The average orientation error can be converged to 0.4 (0.12$\pi$) in cars and 0.7 (0.22$\pi$).

\begin{figure}[tb]
\centering
\subfloat[The loss convergence.]{
\includegraphics[width=0.45\linewidth]{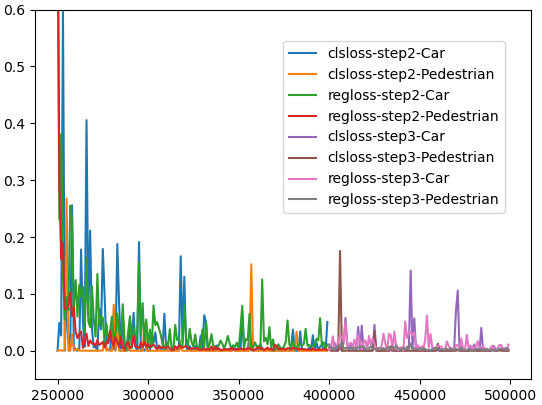}
}
\subfloat[The error decrease.]{
\includegraphics[width=0.45\linewidth]{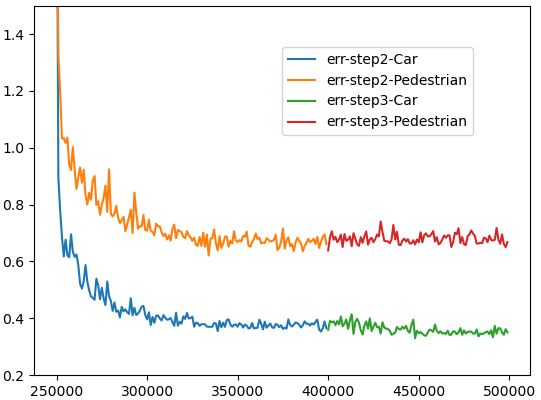}
}
\caption{The loss convergence and error decrease of the joint model in the final step.}
\label{sup_step3}
\end{figure}

In training of the final step, neither of the losses and average errors have large-amplitude changes,
as shown in Fig. \ref{sup_step3}.

\begin{figure}[tb]
\centering
\includegraphics[width=0.6\linewidth]{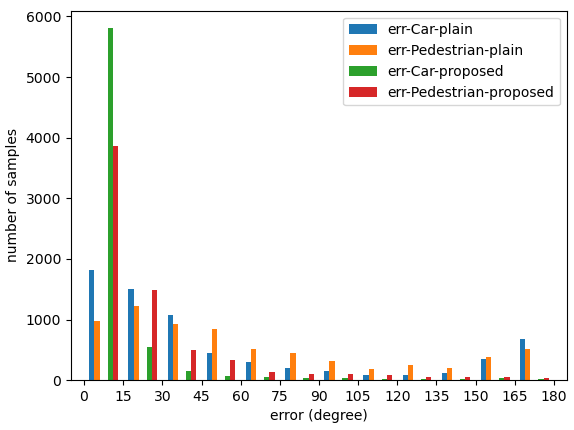}
\caption{MSE comparison of the proposed model and the plain model.}
\label{err}
\end{figure}

Comparison on the mean squared orientation error
of the proposed model and the plain model after identical training is shown in Fig. \ref{err}.
From the comparison, it is obvious that the pretraining method
effectively mitigates the anterior-posterior-similarity problem raised in Section \ref{intro}.

\begin{figure}
\centering
\includegraphics[width=0.95\linewidth]{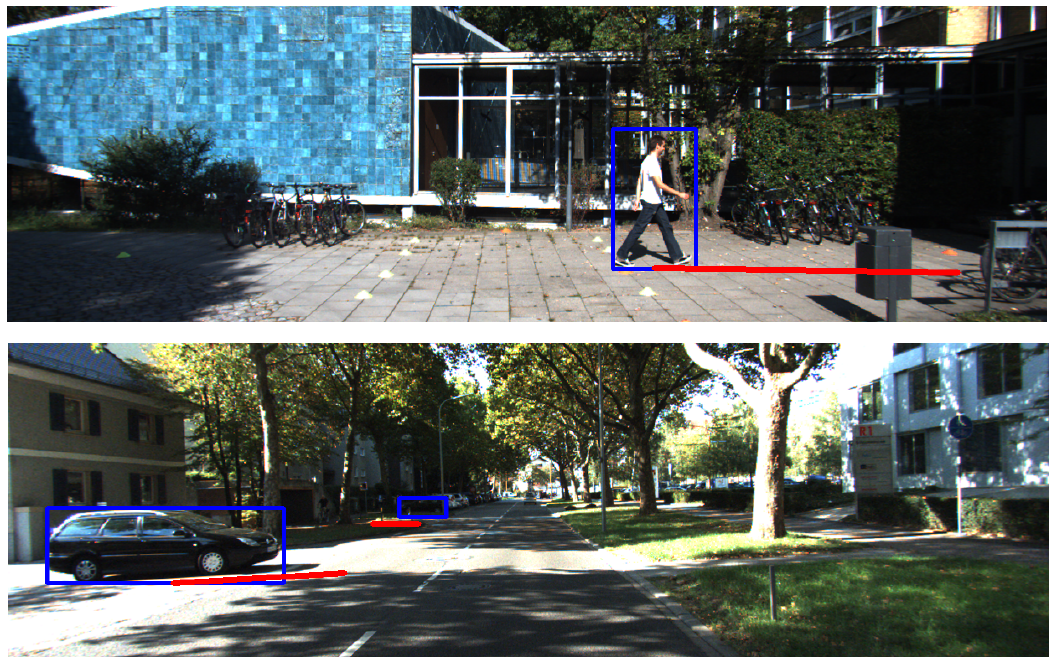}
\caption{Examples of orientation estimation results with full-view monocular images.}
\label{fullview}
\end{figure}

Examples of orientation estimation results with full-view monocular images
are shown in Fig. \ref{fullview}.
Note that in Fig. \ref{fullview}, the 2D object detection results
are the truth labels provided in the dataset.

\subsection{Results of Semisupervised Training on Backbone with Semicircle Classifier}

\begin{figure}[tb]
\centering
\includegraphics[width=0.6\linewidth]{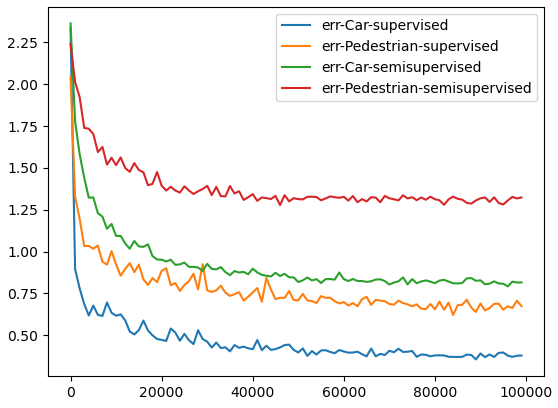}
\caption{The comparison of the supervised and semisupervised methods as the final step.}
\label{semi}
\end{figure}

Fig. \ref{semi} shows the comparison of the supervised and semisupervised methods at the final step.
From the comparison, the semisupervised method fails to train a qualified backbone for orientation estimation.
Possible reasons are listed as follows:
\begin{itemize}
\item Uncertainty of semicircle labels.
In semisupervised training, the labels are set the prediction results of the original images.
Such labels may be wrong enough to mislead the semicircle classifier to wrong semicircle classifications.
\item Oscillations of semisupervised training.
The prediction result of the same data still changes because of update of the network parameters.
Therefore, the labels set in semisupervised training are changed during the process,
which may cause oscillations and even divergence of training.
\end{itemize}

\subsection{Results of the Proposed Method on KITTI Benchmark}

From statements in Section \ref{method}, the proposed method has no ability of 2D object detection.
We use the method proposed in \cite{led} to be the 2D object detector of the proposed method.

The proposed supervised method gets \nth{4} in cars and \nth{2} in pedestrians
among state-of-the-art monocular orientation estimation models
on the KITTI Object Detection and Orientation Estimation Evaluation scoreboard,
with the detailed results shown in Table \ref{kitti}.
AOS on moderate-difficulty-level images is used as metric.
From the baseline comparisons, it is proved that
the proposed method makes a simple backbone reach the state-of-the-art performance in orientation estimation.

\begin{table}[tb]
\centering
\caption{Comparisons between the proposed method and state-of-the-art baselines.}
\begin{scriptsize}
\begin{tabular}{|c|c|c|c|c|c|c|c|}
\hline
\multirow{2}*{Name} & AOS & Rank & AOS & Rank & \multirow{2}*{Time} & 3D \\
 & (car) & (car) & (ped) & (ped) & & info \\
\hline
Mono3D \cite{mono3d} & 87.28\% & \nth{8} & 58.66\% & \nth{3} & 4.2s & \checkmark \\
\hline
Deep3DBox \cite{deep3dbox} & 89.88\% & \nth{4} & \diagbox{}{} & \diagbox{}{} & 1.5s & \checkmark \\
\hline
SubCNN \cite{subcnn} & 89.53\% & \nth{5} & 66.70\% & \nth{1} & 2s & \checkmark \\
\hline
MVRA \cite{mvra} & 94.46\% & \nth{1} & \diagbox{}{} & \diagbox{}{} & 0.18s & \checkmark \\
\hline
Deep MANTA \cite{deepmanta} & 93.31\% & \nth{2} & \diagbox{}{} & \diagbox{}{} & 0.7s & \checkmark \\
\hline
MonoPSR \cite{monopsr} & 87.45\% & \nth{7} & 54.65\% & \nth{4} & 0.2s & \checkmark \\
\hline
Shift R-CNN \cite{shiftrcnn} & 87.47\% & \nth{6} & 46.56\% & \nth{5} & 0.25s & \checkmark \\
\hline
\textbf{Ours(Supervised)} & 90.63\% & \nth{3} & 61.70\% & \nth{2} & 0.23s & \\
\hline
\end{tabular}
\end{scriptsize}
\label{kitti}
\end{table}

\subsection{Discussions}

In this section, two main parts are discussed:
first, the core advantages of the proposed method over baselines;
second, the effectiveness of the proposed method from a hypothetical perspective.

From Table \ref{kitti}, one major difference between the proposed method and baselines is that
the proposed method does not rely on 3D information of the environment in training.
In terms of applications, monocular cameras are more suitable in occasions without 3D sensors,
as stated in Section \ref{intro}.
Moreover, with 3D information, there are better ways with better devices to make intention predictions,
such as LiDAR-based object tracking.
Therefore, our monocular method is more suitable in monocular occasions.

The proposed method is also less complex than most baselines.
The semicircle classifier can be integrated into other orientation estimation models,
indicating that it is a universal method not repelling to baselines.
In contrast, the baselines for comparison mostly require special designs
(such as well-designed 3D object detection head) or additional materials
(such as 3D labels and premade 3D models).
This is another advantage of the proposed method.

The state-of-the-art performance with a simple network structure
is achieved by more precise network training caused by semicircle prediction.

\begin{figure}
\centering
\subfloat[The plain well-trained model.]{
\includegraphics[width=0.3\linewidth]{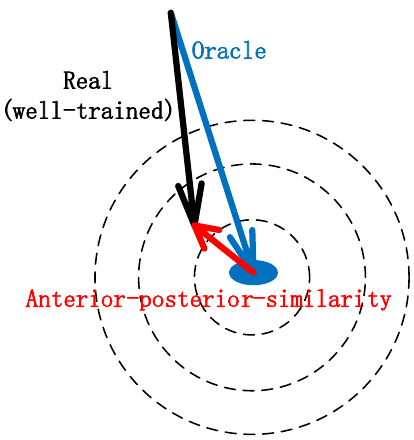}
\label{manifold}
}
\subfloat[The proposed supervised model.]{
\includegraphics[width=0.3\linewidth]{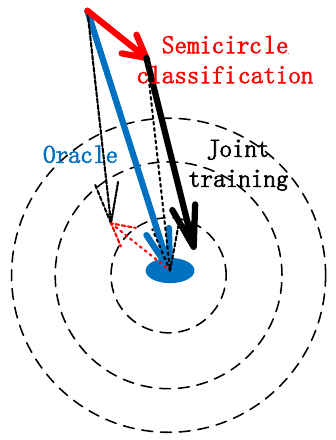}
\label{manifold-semicircle}
}
\subfloat[The proposed semisupervised model.]{
\includegraphics[width=0.3\linewidth]{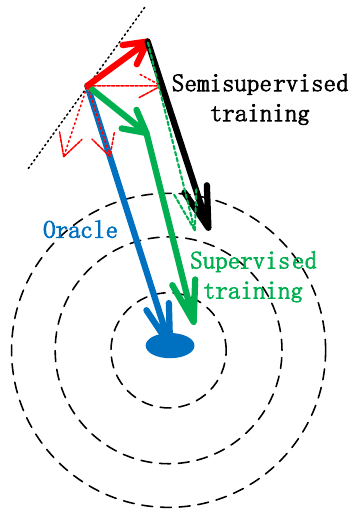}
\label{manifold-semi}
}
\caption{Manifold-based iteration hypothesis of the proposed and the plain model.}
\end{figure}

A manifold of optimization of orientation estimation is shown in Fig. \ref{manifold}.
An ideal model (i.e. oracle) can reach optimal with the ideal iteration direction.
A well-trained model may fail to reach optimal because of anterior-posterior-similarity,
shown in Fig. \ref{manifold} that the iteration direction is different from the oracle.
To be more accurate, the model should continue the iterations until it covers this difference.
The rest of the iterations should be guided with a criterion closely related to the similarity problem,
which, in this paper, is the semicircle classification loss.

With the semicircle classifier, such iterations are completed at the beginning of the training process
(i.e. as pretraining), with the whole iteration process shown in Fig. \ref{manifold-semicircle}.
With the semicircle classifier, the anterior-posterior-similarity problem is mitigated,
and the model reaches better performance.
Due to the differences of models, the iteration direction of the proposed model in joint training
is slightly different from the original plain model.

However, with the semicircle label unknown and `randomly' decided,
the iteration direction in semisupervised training is different from that in supervised training.
The supervised criterion is a stronger restriction to the network than the semisupervised criterion,
which is presented in \eqref{restriction}.
\begin{equation}
\begin{aligned}
&(C(F(x_1))=y) \cap (C(F(x_2))=\overline{y}) \\
&\to C(F(x_1))=\overline{C(F(x_2))}
\end{aligned}
\label{restriction}
\end{equation}
Therefore, any direction can be the iteration direction of semisupervised training
if it has a component with the same direction as the supervised pretraining direction
(the red arrow in Fig. \ref{manifold-semicircle}):
\begin{equation}
\begin{aligned}
\bm{d}_{supervised} \cdot \bm{d}_{semisupervised} > 0
\end{aligned}
\end{equation}

An instance is shown in Fig. \ref{manifold-semi},
with the semisupervised pretraining direction
satisfying all conditions but having huge difference from the supervised direction.
This may severely reduce the performance of the joint model,
since the flawed classifier acts as a misleading factor of the feature extractor in joint training.
The semisupervised model in this paper is an example of, but not limited to, this instance.

\section{Conclusion}\label{conclusion}

We propose a pretraining method to enhance the robustness of orientation estimation models.
We pretrain the model to classify the semicircle in which the orientation angle is located,
to develop its ability of classifying the similar anterior and posterior parts of objects
and extracting basic orientation features in the image.
The supervised and semisupervised versions of the method are both proposed, analyzed and experimented.
Experiments show that the pretraining step
contributes to the accuracy and robustness of orientation estimation.
The proposed method enhances the accuracy of orientation estimation models,
and mitigates the serious safety threats in autonomous driving to a certain extent.

\bibliographystyle{unsrt}
\bibliography{root}

\end{document}